\documentclass{article} 
\usepackage{iclr2027_conference,times}

\usepackage{amsmath,amsfonts,bm}

\def\eqref#1{equation~\ref{#1}}

\def\1{\bm{1}}

\DeclareMathAlphabet{\mathsfit}{\encodingdefault}{\sfdefault}{m}{sl}
\SetMathAlphabet{\mathsfit}{bold}{\encodingdefault}{\sfdefault}{bx}{n}

\usepackage{hyperref}
\usepackage{url}
\usepackage{graphicx}
\usepackage{caption}
\usepackage{booktabs}
\usepackage[table]{xcolor}
\usepackage{multirow}
\usepackage{subcaption}

\title{A Lightweight Velocity Correction Framework for Improving Instruction Following in Few-Step Text-to-Image Generation}

\author{
Yating Gao$^{1,2}$ \quad
Zhenlong Zhu$^{2}$ \quad
Wen Zihao$^{2,3}$ \quad
Lenan Zhang$^{4,2}$ \\
Jiexiong Wang$^{2}$ \quad
Yang Yang$^{2}$ \quad
Guanjun Jiang$^{2}$ \\
Yixiong Zou$^{4}$ \quad
Takuya Maekawa$^{1}$ \\[0.6em]
$^{1}$The University of Osaka\quad
$^{2}$Alibaba Group \\
$^{3}$Peking University \\
$^{4}$Huazhong University of Science and Technology
}

\iclrfinalcopy 
\begin{document}

\maketitle
\suppressfloats[t]

\begin{abstract}
Few-step text-to-image generators produce high-quality images with few network forward passes, yet struggle with complex prompts involving multiple subjects, interactions, spatial relations, counting, and attribute binding.
Can these errors be reduced without changing the generator's parameters or increasing its forward-pass count?
We propose a lightweight correction framework for a frozen 6B-parameter generator.
The generator predicts velocities that determine how its intermediate image representation is updated; a corrector with fewer than 20M parameters learns residual adjustments at selected sampling steps.
Because earlier corrections change the states encountered at later steps, our On-Trajectory Flow-Bridge Supervision derives residual targets at the states actually reached during corrected generation.
A single corrector is shared across selected steps and runs in parallel with the generator.
Only the corrector is trained, requiring neither reward optimization nor re-distillation.
To our knowledge, this is the first framework for lightweight velocity-residual adaptation of fully frozen few-step text-to-image generators whose corrector runs in parallel with the generator while preserving its original sampling schedule and forward-pass count.
On our target complex-scene distribution, the framework achieves a NetWin of +17.60 percentage points over frozen Z-Image-Turbo in pairwise instruction-following evaluation, with only 26.9 ms (0.84\%) additional end-to-end latency per image and no increase in backbone evaluations.
\end{abstract}

\section{Introduction}
\label{sec:introduction}

Few-step text-to-image generation aims to recover the effectiveness of multi-step denoising with only a limited number of network forward passes. Recent diffusion and flow distillation methods have made substantial progress toward this goal \citep{kim2024ctm,yin2024dmd2,ren2024hypersd,zheng2026rcm}. Models such as Z-Image-Turbo already produce images of high visual quality \citep{zimage2025,jiang2025dmdr}, yet their adherence to complex instructions remains limited, particularly for multi-subject interactions, spatial relations, counting, and attribute binding. Thus, visually plausible outputs can still miss important details specified in the prompt.

\begin{figure}[t]
    \centering
    \includegraphics[width=\linewidth]{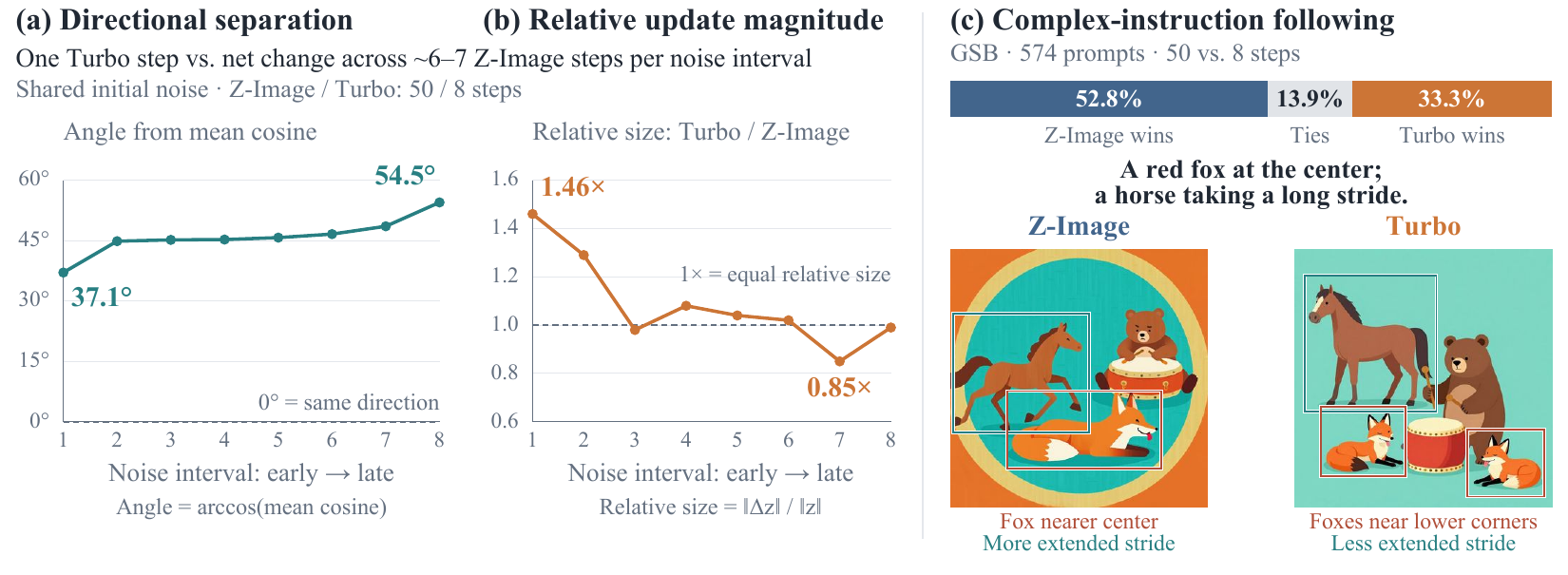}
    \caption{\textbf{Few-step and multi-step generation: intermediate updates and instruction following.}
    \textbf{(a)} Intermediate update directions differ substantially, with angles corresponding to mean cosine similarities increasing from $37.1^\circ$ to $54.5^\circ$.
    \textbf{(b)} Relative update magnitudes remain broadly comparable, with Turbo-to-Z-Image ratios ranging from $0.85$ to $1.46$.
    \textbf{(c)} On 574 complex-scene prompts, 50-step Z-Image outperforms eight-step Z-Image-Turbo by +19.5 pp NetWin in pairwise instruction-following evaluation. The example highlights differences in satisfying the specified fox placement and horse stride.}
    \label{fig:velocity_effect}
\end{figure}

Existing post-training methods improve instruction following by adapting the generator \citep{jiang2024comat,black2024ddpo,prabhudesai2024alignprop,jiang2025dmdr,jiang2026dopsd}. However, adapting a step-distilled generator can disrupt its learned few-step sampling behavior, compromising image quality under the original schedule. To explore an alternative, we draw inspiration from the stronger complex-instruction following of 50-step Z-Image compared with eight-step Z-Image-Turbo on our target distribution (Fig.~\ref{fig:velocity_effect}(c)). Comparing their sampling trajectories from shared initial noise reveals substantial differences in update direction over matched noise intervals, despite broadly comparable relative update magnitudes (Fig.~\ref{fig:velocity_effect}(a,b)). This suggests intermediate velocities as a potential intervention point: can we correct these velocities to improve few-step instruction following while keeping the generator frozen and its sampling schedule unchanged?

\begin{figure}[t]
    \centering
    \includegraphics[width=\linewidth]{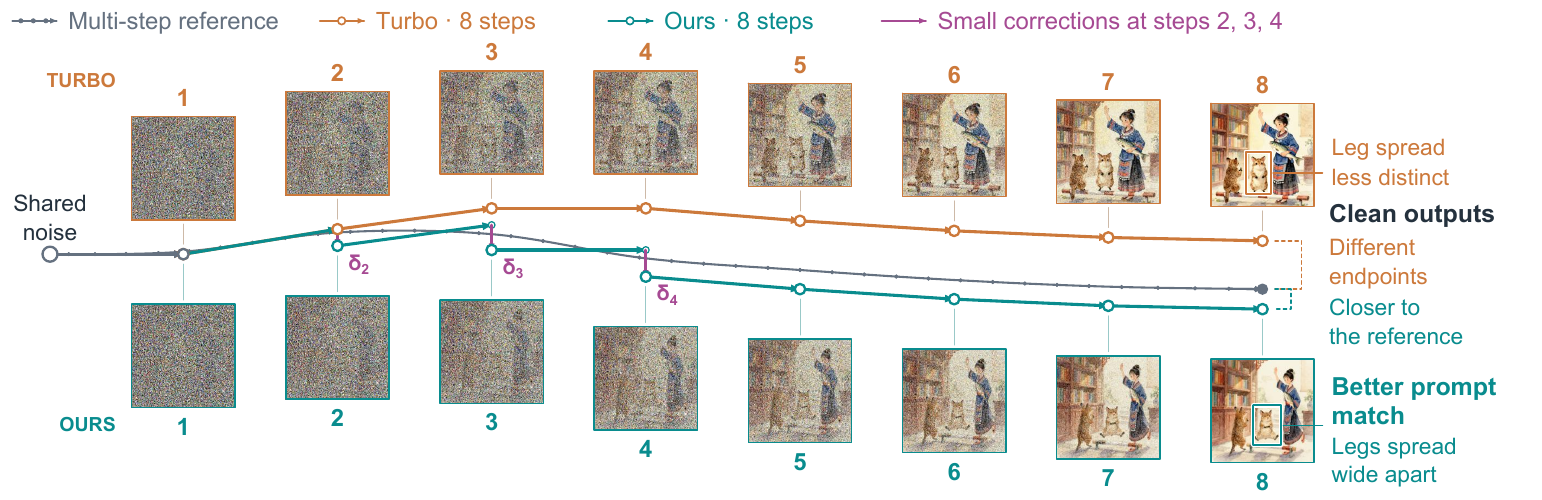}
    \caption{\textbf{Sparse velocity correction for few-step generation.} From shared noise, Turbo (orange) and Ours (teal) generate images in eight steps, alongside a multi-step reference (gray). Ours freezes the Turbo backbone and adds velocity residuals $r_{\phi,k}$ only at steps $k\in\{2,3,4\}$. Purple arrows $\delta_k$ indicate the resulting trajectory displacements; markers link intermediate states to image thumbnails. The highlighted animal in Ours better satisfies ``its front and hind legs spread wide apart.'' Trajectories and endpoint gaps are schematic.}
    \label{fig:velocity_residual_correction}
    \vspace{-5mm}
\end{figure}

Our key idea is to keep the generator frozen and train a lightweight auxiliary branch to add residual corrections to its predicted latent-space velocities, preserving the sampling schedule and backbone forward-pass count. A step-distilled generator has learned update behavior adapted to its own sampling schedule, so directly matching a multi-step model's intermediate velocities may not preserve its few-step denoising behavior. We therefore learn residual corrections from target-image supervision along the corrected few-step trajectories. Their effectiveness depends on both prediction accuracy and intervention timing. When a correction is insufficiently accurate, applying it at a late step may leave too few subsequent updates to resolve the resulting deviation. We therefore apply corrections at selected early steps, allowing the remaining frozen updates to accommodate residual inaccuracies and complete denoising.

Realizing this idea requires addressing three challenges: limiting inference overhead, defining suitable residual targets, and learning sufficiently accurate corrections. \textbf{Parallel Velocity Residual Correction} addresses the first by conditioning on the current latent and cached previous-step velocity, allowing the corrector to run alongside the current generator evaluation. To define suitable targets, \textbf{On-Trajectory Flow-Bridge Supervision} constructs a linear reference bridge from the initial noise to a target latent obtained by offline editing of the generator's output. Because earlier corrections change the states encountered at later steps, it recomputes residual targets at the states actually visited by the corrected sampler, combining reference-velocity correction with trajectory recovery. To improve correction accuracy, \textbf{Shared Multi-Step Learning} jointly trains a single noise-conditioned corrector across multiple selected early steps using local supervision, without backpropagating through preceding sampling updates. Figure~\ref{fig:velocity_residual_correction} illustrates the resulting improvement within the original eight-step schedule: the animal's pose better satisfies ``its front and hind legs spread wide apart.''

Our main contributions are:
\par\smallskip
\noindent\textbullet\ We introduce a lightweight external velocity-correction framework for fully frozen few-step text-to-image generators. To our knowledge, it is the first to learn velocity residuals with a corrector running in parallel with the generator while preserving its parameters, sampling schedule, and backbone forward-pass count.
\par\smallskip
\noindent\textbullet\ We empirically characterize how correction timing affects few-step generation. These findings motivate applying corrections at selected early steps, leaving subsequent denoising updates available to accommodate inaccuracies in the learned residuals.
\par\smallskip
\noindent\textbullet\ We develop On-Trajectory Flow-Bridge Supervision for state-dependent residual targets combining reference-velocity correction with trajectory recovery, and Shared Multi-Step Learning for training a single corrector across selected steps with local supervision.
\par\smallskip
\noindent\textbullet\ On our target complex-scene distribution, an 18.6M-parameter corrector achieves +17.60 pp NetWin over frozen Z-Image-Turbo, with only 0.84\% additional inference latency and 0.04 GiB additional peak allocated GPU memory.
\par

\section{Related Work}
\label{sec:related_work}

\textbf{Fine-Tuning for Text-to-Image Generation.} Supervised and feedback-based fine-tuning improves text--image alignment: CoMat addresses concept omission and attribute binding, while CRAFT and DreamSync use curated or feedback-filtered supervision \citep{jiang2024comat,sun2026craft,sun2023dreamsync}. LoRA supports parameter-efficient adaptation \citep{hu2022lora}. For few-step generators, fine-tuning can disrupt sampling behavior learned through specialized distillation objectives, including LCM, TCD, Hyper-SD, DMD2, and SDXL-Lightning \citep{luo2023lcm,zheng2024tcd,ren2024hypersd,yin2024dmd2,lin2024sdxllightning}. Our method keeps the generator frozen and trains an auxiliary velocity corrector.

\textbf{Reward- and Preference-Based Alignment.} ImageReward, PickScore, and HPSv2 provide learned preference signals \citep{xu2023imagereward,kirstain2023pickapic,wu2023hpsv2}, while DDPO, AlignProp, and DRaFT optimize generators through reward feedback \citep{black2024ddpo,prabhudesai2024alignprop,clark2024draft}. Diffusion-DPO, D3PO, and Diffusion-RPO support preference optimization without separately trained reward models \citep{wallace2024diffusiondpo,yang2024d3po,gu2024diffusionrpo}, with further developments in DiffusionNFT, TreeGRPO, and DiffusionOPSD \citep{zheng2026diffusionnft,ding2026treegrpo,zhou2026diffusionopsd}. Multiple compositional requirements and evaluator reliability complicate reward design \citep{chen2025mjbench,li2026qwenimagebench}. Although generated and target images could also form preference pairs for Diffusion-DPO, we use target images to derive explicit stepwise velocity-residual supervision at corrected sampler states.

\textbf{Trajectory-Based Adaptation and Control.} ControlNet and T2I-Adapter add conditioning modules to frozen diffusion models \citep{zhang2023controlnet,mou2023t2iadapter}; Universal Guidance and FreeControl provide training-free control \citep{bansal2024universalguidance,mo2024freecontrol}; and D-Flow, FlowChef, and VESFlow intervene in flow trajectories or vector fields \citep{benhamu2024dflow,patel2024flowchef,choi2026vesflow}.

D-OPSD adapts step-distilled generators by matching target-image-informed teacher predictions on student rollouts \citep{jiang2026dopsd}. We instead derive residual targets from the target latent and current frozen-generator prediction, without a teacher branch or generator updates.

Residual Learning corrects samples near the end of diffusion sampling \citep{zhang2024residual}. Diffusion Controller conditions its side network on the reverse-transition mean from the current backbone prediction \citep{yang2026diffcon}, introducing a within-step dependency. Our corrector uses the current latent and cached preceding-step velocity, enabling parallel execution with the current backbone evaluation.

\section{Method}
\label{sec:method}

\begin{figure*}[t]
    \centering
    \includegraphics[width=\textwidth]{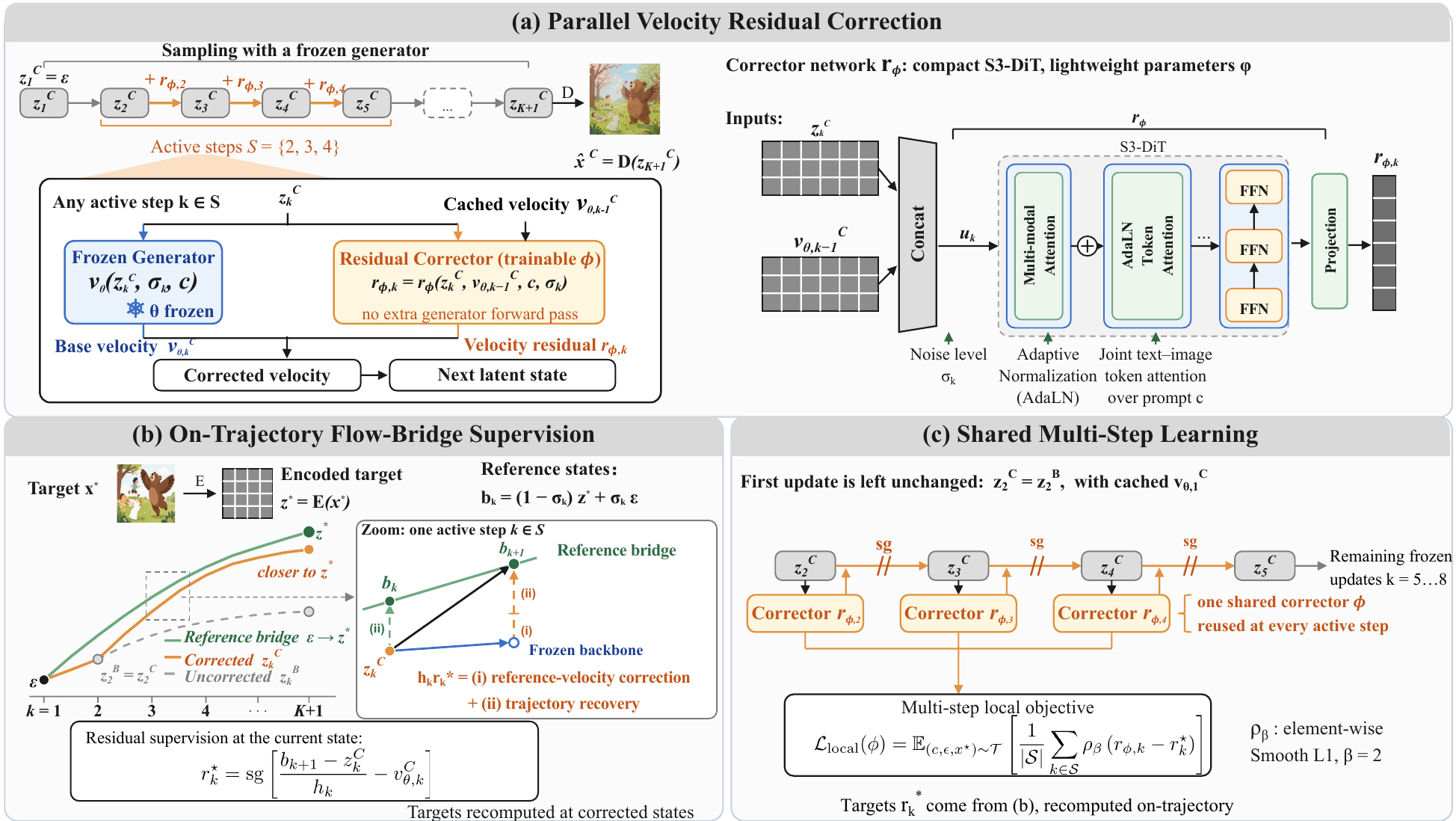}
    \caption{\textbf{Overview of velocity residual correction.}
    A lightweight corrector augments a frozen few-step generator without changing its sampling schedule or forward-pass count.
    \textbf{(a) Parallel Velocity Residual Correction.}
    Using the current latent and cached previous-step velocity, the corrector runs alongside the generator and adds a residual to its predicted velocity.
    \textbf{(b) On-Trajectory Flow-Bridge Supervision.}
    A noise-to-target reference bridge defines residual targets at corrected states, combining reference-velocity correction with trajectory recovery.
    \textbf{(c) Shared Multi-Step Learning.}
    One corrector is trained across selected steps with local supervision and no backpropagation through sampling updates. $r_{\phi,k}$ and $r_k^\star$ denote predicted and target residuals.}
    \label{fig:method_overview}
\end{figure*}

As shown in Fig.~\ref{fig:method_overview}, our framework adds velocity residuals at selected sampling steps of a frozen generator, preserving its sampling schedule and backbone forward-pass count.

\textbf{Problem Setup.} We consider few-step generators whose sampling updates admit an Euler-form velocity parameterization with a fixed noise schedule
$\sigma_1=1>\sigma_2>\cdots>\sigma_K>\sigma_{K+1}=0$,
where $K$ is the number of sampling updates.
We instantiate the framework with Z-Image-Turbo using $K=8$.
Let $z_k$ denote the latent state before update $k$, $c$ the text condition, and
$v_{\theta}(z_k,\sigma_k,c)$ the latent-space velocity predicted by the pretrained generator with frozen parameters $\theta$.
This velocity specifies how the latent state moves along the generation trajectory as the noise level changes.

The original sampling update is
\begin{equation}
z_{k+1}^{B}
=
z_k^{B}
+
h_k v_{\theta}(z_k^{B},\sigma_k,c),
\qquad
h_k=\sigma_{k+1}-\sigma_k<0,
\label{eq:base_transition}
\end{equation}
where $B$ denotes the uncorrected trajectory and $h_k<0$ is the signed sampling step size as the noise level decreases.
Starting from $z_1^{B}=\epsilon$, with
$\epsilon\sim\mathcal{N}(0,I)$,
the sampler produces the final latent $z_{K+1}^{B}$, which is decoded as
$\hat{x}^{B}=\mathcal{D}(z_{K+1}^{B})$
by the pretrained decoder $\mathcal{D}$.
Throughout, $\mathcal{E}$ includes the backbone-specific latent normalization used by the pretrained generator, and $\mathcal{D}$ includes its inverse before VAE decoding. All states and bridge targets are expressed in this common coordinate system.

\subsection{Parallel Velocity Residual Correction}
\label{sec:parallel_residual}

Starting from $z_1^C=\epsilon$, we apply a corrector $r_\phi$ at
steps $\mathcal{S}\subseteq\{2,\ldots,K\}$, leaving the first update
unchanged. The frozen generator predicts
$v_{\theta,k}^C=v_\theta(z_k^C,\sigma_k,c)$.
At each active step, the corrector takes the current latent and
cached preceding-step generator velocity as inputs:
\begin{equation}
r_{\phi,k}
=
r_{\phi}
\left(
z_k^{C},
v_{\theta,k-1}^{C},
c,
\sigma_k
\right),
\qquad
v_{\theta,k-1}^{C}
=
v_{\theta}(z_{k-1}^{C},\sigma_{k-1},c).
\label{eq:residual_corrector}
\end{equation}

We instantiate $r_{\phi}$ with a compact S3-DiT architecture following ~\cite{zimage2025}.
It takes the channel-wise concatenation
$u_k=\operatorname{Concat}[z_k^{C};v_{\theta,k-1}^{C}]$
as its latent input and incorporates $\sigma_k$ through adaptive normalization.
The corrector reuses the precomputed text token embeddings from the frozen generator's text encoder, requiring no additional text-encoding forward pass. These embeddings are jointly processed with the corrector's image tokens, enabling token-level text--image interactions.

The corrected velocity and sampling update are
\begin{equation}
\hat{v}_k
=
\begin{cases}
v_{\theta,k}^{C}+r_{\phi,k}, & k\in\mathcal{S},\\
v_{\theta,k}^{C}, & k\notin\mathcal{S},
\end{cases}
\qquad
z_{k+1}^{C}
=
z_k^{C}+h_k\hat{v}_k.
\label{eq:corrected_update}
\end{equation}

For a fixed generator, the current velocity is a deterministic function of $(z_k^{C},\sigma_k,c)$, all available to the corrector; the cached previous-step velocity supplies an additional trajectory-dependent feature. The corrector thus learns the residual without taking the current generator output as an input, allowing both branches to execute concurrently. The current output is used to construct the training target in Sec.~\ref{sec:flow_bridge} and combined with the predicted residual in Eq.~(\ref{eq:corrected_update}).

\subsection{On-Trajectory Flow-Bridge Supervision}
\label{sec:flow_bridge}

\begin{figure}[t]
    \centering
    \includegraphics[width=0.7\linewidth]{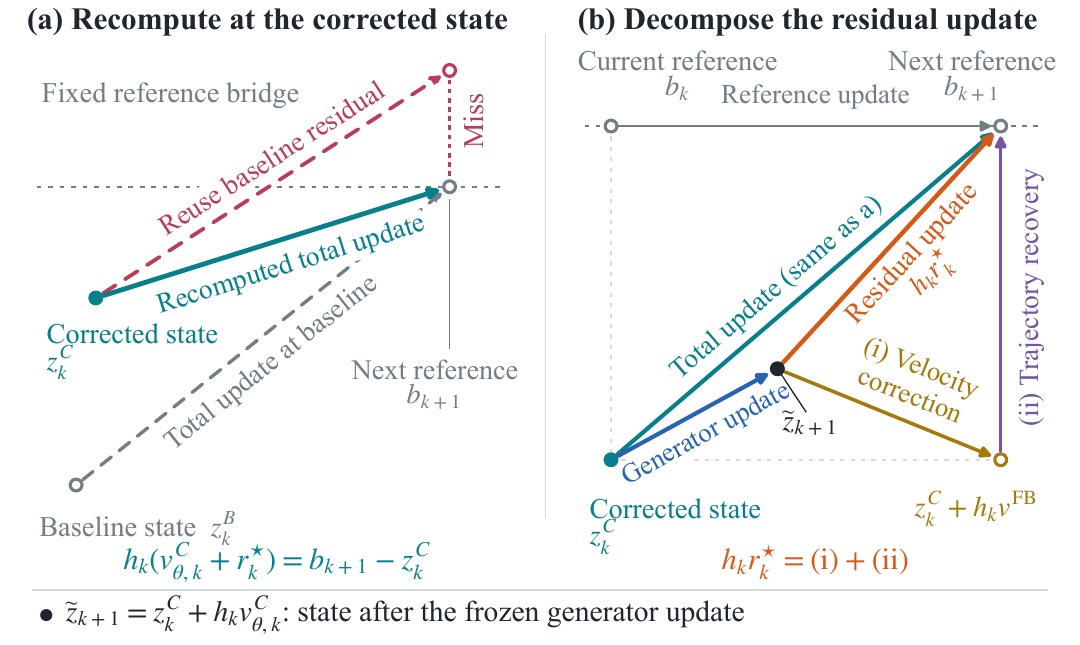}
    \caption{\textbf{On-Trajectory Flow-Bridge Supervision.} \textbf{(a)} Reusing a residual target from $z_k^B$ at $z_k^C$ would cause a mismatch (red dashed). We instead recompute targets at states visited by the corrected sampler during training. \textbf{(b)} The total update (teal in both panels) sums the generator update (blue) and residual update (orange), which decomposes into reference-velocity correction (gold) and trajectory recovery (purple). The black dot denotes the generator-only next state
$\widetilde{z}_{k+1}=z_k^C+h_kv_{\theta,k}^C$. Arrows denote state displacements.}
    \label{fig:flow_bridge_mechanism}
    \vspace{-3mm}
\end{figure}

During training, corrected states $z_k^{C}$ are generated online by rolling out the sampler with the current corrector, rather than being precomputed from the baseline trajectory. Before the first active correction, $z_2^{C}=z_2^{B}$; later corrected states are produced by preceding corrected updates. Therefore, a residual target computed at the baseline state $z_k^{B}$ may no longer be valid at the actually visited state $z_k^{C}$. As illustrated in Fig.~\ref{fig:flow_bridge_mechanism}(a), we instead recompute the required velocity correction at each corrected state using a fixed reference path from the initial noise to the target latent.

Given the target image $x^{\star}$, we obtain its clean latent $z^{\star}=\mathcal{E}(x^{\star})$ using the pretrained encoder $\mathcal{E}$. With the initial noise $\epsilon$ used to generate the original image, we define
\begin{equation}
b_k
=
(1-\sigma_k)z^{\star}
+
\sigma_k\epsilon.
\label{eq:bridge_state}
\end{equation}
This linear reference path connects $b_1=\epsilon$ to $b_{K+1}=z^{\star}$ and need not follow the generator's trajectory. It remains fixed for each training example, while residual targets are recomputed at the changing corrected states.

At each active step, the online rollout provides the current corrected state $z_k^{C}$, at which we evaluate the frozen generator to obtain $v_{\theta,k}^{C}$. The velocity required to reach $b_{k+1}$ in one Euler update is $(b_{k+1}-z_k^{C})/h_k$. Subtracting the generator prediction gives the residual target
\begin{equation}
r_k^{\star}
=
\operatorname{sg}
\left[
\frac{b_{k+1}-z_k^{C}}{h_k}
-
v_{\theta,k}^{C}
\right],
\label{eq:bridge_residual}
\end{equation}
where $\operatorname{sg}[\cdot]$ denotes stop-gradient.

To explain how this target accounts for both the generator prediction and the current state, we decompose it as
\begin{equation}
r_k^{\star}
=
\operatorname{sg}
\left[
\underbrace{
v^{\mathrm{FB}}-v_{\theta,k}^{C}
}_{\text{reference-velocity correction}}
+
\underbrace{
\frac{b_k-z_k^{C}}{h_k}
}_{\text{trajectory recovery}}
\right],
\qquad
v^{\mathrm{FB}}
=
\frac{b_{k+1}-b_k}{h_k}
=
\epsilon-z^{\star}.
\label{eq:bridge_decomposition}
\end{equation}
The first term corrects the discrepancy between the reference velocity and the generator prediction, while the second compensates for displacement of the current state from the reference bridge. Figure~\ref{fig:flow_bridge_mechanism}(b) illustrates these terms as state displacements: multiplying them by $h_k$ gives the reference-velocity correction and trajectory-recovery arrows, whose sum is the required residual update $h_k r_k^{\star}$.

For a fixed current state, the residual prediction error directly determines the next-state bridge error:
\begin{equation}
z_{k+1}^{C}-b_{k+1}
=
h_k\left(r_{\phi,k}-r_k^{\star}\right),
\qquad k\in\mathcal{S}.
\label{eq:residual_state_error}
\end{equation}
An exact residual prediction therefore reaches $b_{k+1}$ in one step, but subsequent uncorrected updates need not remain on the bridge.
During training, we roll out the corrected sampler and recompute the residual target in Eq.~(\ref{eq:bridge_residual}) at every active step.
\subsection{Shared Multi-Step Learning}
\label{sec:shared_multistep}

We share a single noise-conditioned corrector across selected steps and train it with local supervision. Eq.~(\ref{eq:residual_state_error}) shows that each visited state provides a supervised training instance without differentiating through the preceding sampling updates.

We zero-initialize the corrector's output projection so that training starts from the original generator trajectory. Our default active-step set is $\mathcal{S}=\{2,3,4\}$, with the previous-step generator velocity cached as input to the corrector.

For paired training tuples $(c,\epsilon,x^{\star})\sim\mathcal{T}$, we minimize
\begin{equation}
\mathcal{L}_{\mathrm{local}}(\phi)
=
\mathbb{E}_{(c,\epsilon,x^{\star})\sim\mathcal{T}}
\left[
\frac{1}{|\mathcal{S}|}
\sum_{k\in\mathcal{S}}
\rho_{\beta}\left(r_{\phi,k}-r_k^{\star}\right)
\right],
\label{eq:local_loss}
\end{equation}
where $\mathcal{T}$ denotes the training set and $\rho_{\beta}$ is the Smooth L1 loss averaged over latent elements. We use $\beta=2$ by default and weight active steps equally in velocity-residual space.

We detach rollout states and cached velocities between steps and treat the generator predictions and residual targets as constants. Thus, later losses are evaluated on states produced by earlier corrections without backpropagating through the preceding corrections or sampling updates.

The pretrained generator, text encoder, and latent encoder--decoder remain frozen throughout training. At inference, the shared corrector is applied at the same active steps, and the generator completes the remaining updates before decoding $\hat{x}^{C}=\mathcal{D}(z_{K+1}^{C})$.

\section{Experiments}
\label{sec:experiments}

\subsection{Experimental Setup}
\label{sec:experimental_setup}

\textbf{Data.}
We use approximately 4K compositional captions with a 90/10 training/validation split.
Offline target images are generated by editing frozen-backbone outputs with GPT-Image-2~\citep{openai_gpt_image2}.
Each target is paired with the original caption and the initial noise used to generate the backbone output.
Evaluation uses 574 held-out prompts with no normalized exact overlap with either the training or validation prompts (Appendix~\ref{app:data_details}).

\textbf{Implementation and baselines.} We equip frozen eight-step Z-Image-Turbo~\citep{zimage2025} with an 18.6M-parameter S3-DiT corrector shared across $\mathcal{S}=\{2,3,4\}$. Only the corrector is trained, using AdamW with learning rate $2\times10^{-4}$ and Smooth L1 loss ($\beta=2$), without backpropagation through rollout updates; the backbone, text encoder, and VAE remain frozen. We compare against frozen Turbo, SFT+LoRA~\citep{hu2022lora}, DPO+LoRA~\citep{wallace2024diffusiondpo}, D-OPSD~\citep{jiang2026dopsd}, and DiffCon~\citep{yang2026diffcon}, using the same backbone and velocity parameterization and evaluating all methods with eight sampling steps. SFT+LoRA, DPO+LoRA, and D-OPSD each use 19.85M trainable parameters, while DiffCon uses a 20.21M-parameter controller. Architecture, optimization, and baseline implementation details are provided in Appendix~\ref{app:implementation_details}.

\textbf{Evaluation metrics.}
We conduct Good/Same/Bad (GSB) comparisons against frozen Turbo on 574 held-out target-domain prompts with matched seeds and both image orders. We report $\mathrm{NetWin}=100(G-B)/N$ in percentage points (pp), where $G$, $S$, and $B$ denote adapted-model wins, ties, and baseline wins, respectively, and $N=G+S+B=574$. GSB is used only for evaluation; judging details and repeatability results appear in Appendix~\ref{app:gsb_evaluation_details}. We additionally report category-wise scores on T2I-CompBench++~\citep{huang2025t2icompbenchpp}, using the complete official test set with four images per prompt. Evaluator and sampling details appear in Appendix~\ref{app:compbench_evaluation_details}.

\textbf{Inference efficiency.}
We measure end-to-end latency on one NVIDIA A800 at batch size one, BF16 precision, and $1024\times1024$ resolution. Each method uses the same 50 held-out prompts and three seeds (150 generations), with eight backbone evaluations per image. Timing includes text encoding, denoising, and VAE decoding, but excludes model loading and image saving. We report absolute and relative increases in mean latency across generations, together with the increase in peak allocated GPU memory, all relative to frozen Z-Image-Turbo.


\begin{table*}[t]
    \centering
    \small
    \setlength{\tabcolsep}{4pt}
    \caption{Results on Z-Image-Turbo. NetWin is measured against the frozen baseline on 574 held-out target-domain prompts. All metrics are higher-is-better; best scores are bold, and second-best scores achieved by our method are underlined.}
    \label{tab:main_results}
    \resizebox{\textwidth}{!}{%
    \begin{tabular}{@{}lcccccccccc@{}}
        \toprule
        & & Target-domain & \multicolumn{8}{c}{General-purpose} \\
        \cmidrule(lr){3-3}
        \cmidrule(lr){4-11}
        Method & Trainable & NetWin (pp)
        & Color & Shape & Texture
        & 2D Spatial & 3D Spatial
        & Numeracy & Non-spatial & Complex \\
        \midrule

        Frozen baseline
        & -- & 0.00
        & 0.7919 & 0.5731 & 0.7350
        & 0.3349 & \textbf{0.4158}
        & 0.6738 & 0.3121 & 0.3927 \\

        SFT + LoRA
        & 19.85M & 3.83
        & 0.7962 & 0.5706 & 0.7351
        & 0.3359 & 0.4156
        & 0.6704 & 0.3125 & 0.3947 \\

        DPO + LoRA
        & 19.85M & 1.05
        & 0.7929 & 0.5722 & 0.7347
        & \textbf{0.3375} & 0.4148
        & 0.6757 & 0.3122 & 0.3928 \\

        D-OPSD
        & 19.85M & -12.37
        & 0.7597 & 0.5631 & 0.7206
        & 0.3286 & 0.4147
        & 0.6382 & 0.3109 & 0.3880 \\

        DiffCon adaptation
        & 20.21M & 1.74
        & 0.7827 & 0.5843 & \textbf{0.7495}
        & 0.3305 & 0.4133
        & \textbf{0.6838} & 0.3119 & 0.3938 \\

        \rowcolor{blue!6}
        \textbf{Ours}
        & 18.6M & \textbf{17.60}
        & \textbf{0.8171} & \textbf{0.5887} & \underline{0.7440}
        & \underline{0.3361} & 0.4107
        & 0.6748 & \textbf{0.3132} & \textbf{0.4026} \\
        \bottomrule
    \end{tabular}%
    }
    \vspace{-3mm}
\end{table*}

\begin{figure}[t]
    \centering
    \includegraphics[width=\linewidth]{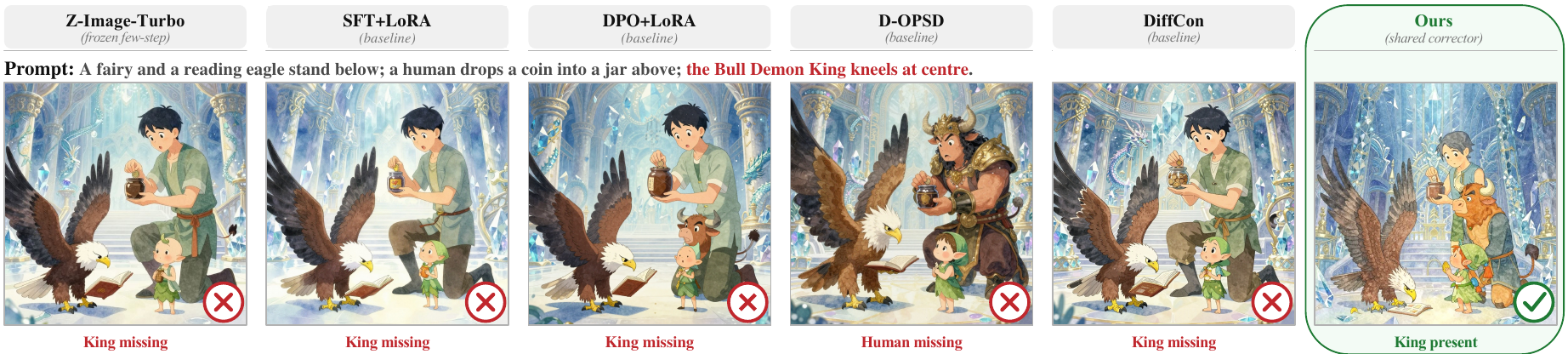}
    \caption{Compositional errors in few-step generation and qualitative comparisons.  Our method better satisfies the highlighted requirements than competing methods.}
    \label{fig:results}
    \vspace{-4.2mm}
\end{figure}

\begin{figure}[t]
    \centering
    \includegraphics[width=\linewidth]{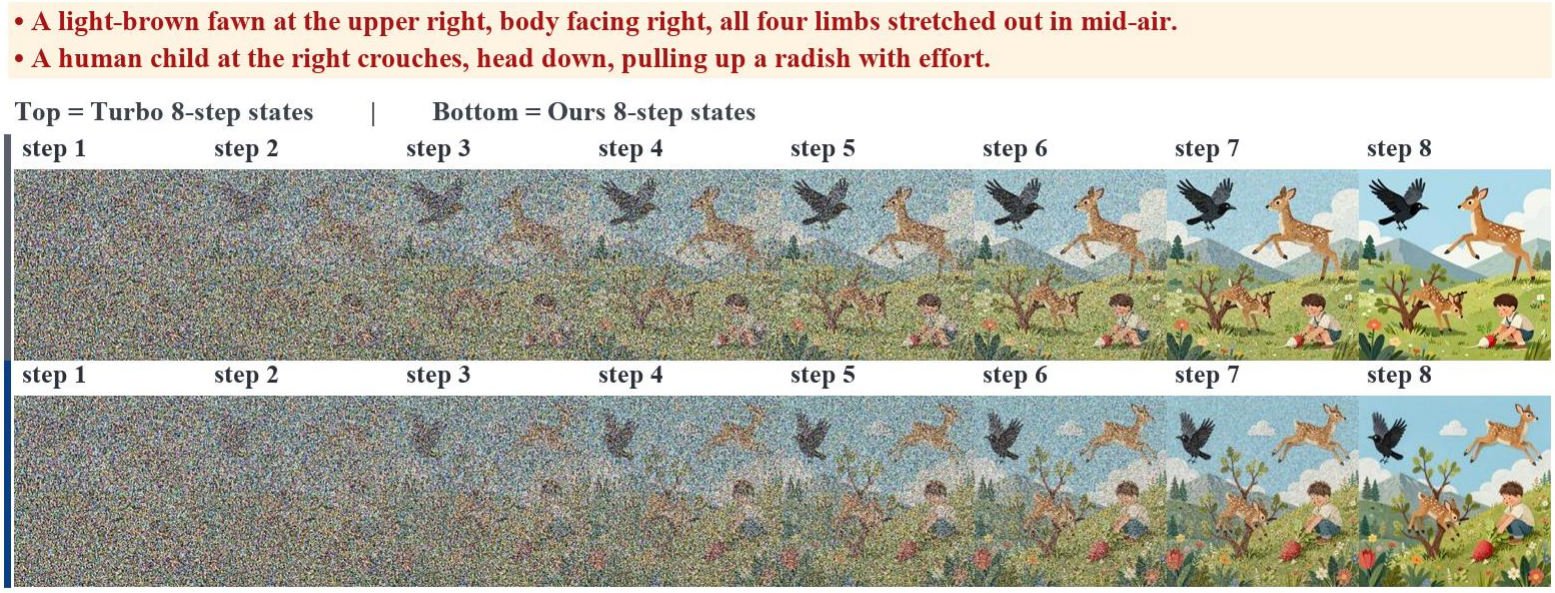}
    \caption{Step-wise qualitative comparison with frozen Z-Image-Turbo. The upper row shows the baseline and the lower row shows our method across eight sampling steps.}
    \label{fig:perstep_comparison}
\end{figure}

\begin{table}[t]
    \centering
    \small
    \setlength{\tabcolsep}{5pt}
    \caption{Target-domain NetWin and inference overhead relative to frozen Z-Image-Turbo. Bold values indicate the best results among adaptation methods.}
    \label{tab:inference_efficiency}
    \begin{tabular}{lrrrr}
        \toprule
        Method
        & \shortstack{NetWin\\(pp)$\uparrow$}
        & \shortstack{Latency\\overhead (\%)$\downarrow$}
        & \shortstack{Added latency\\(ms)$\downarrow$}
        & \shortstack{Memory increase\\(GiB)$\downarrow$} \\
        \midrule
        Frozen Turbo
        & 0.00 & 0.00 & 0.0 & 0.00 \\
        DiffCon
        & +1.74 & 3.24 & 103.9 & \textbf{0.02} \\
        \rowcolor{blue!6}
        Ours
        & \textbf{+17.60} & \textbf{0.84} & \textbf{26.9} & 0.04 \\
        \bottomrule
    \end{tabular}
    \vspace{-3mm}
\end{table}

\subsection{Main Results}
\label{sec:main_results}

Our 18.6M-parameter corrector achieves the highest target-domain NetWin (+17.60 pp) among the evaluated methods (Table~\ref{tab:main_results}). On T2I-CompBench++, it leads in four categories and ranks second in two, improving seven of eight scores over frozen Z-Image-Turbo with a small decrease in 3D spatial relations. Figure~\ref{fig:results} illustrates improved adherence to compositional requirements relative to the evaluated baselines.

Under the unchanged eight-step schedule, our method adds only 0.84\% end-to-end latency (26.9 ms/image) and 0.04 GiB peak allocated GPU memory, achieving higher target-domain NetWin with lower latency overhead than the auxiliary-network baseline DiffCon (Table~\ref{tab:inference_efficiency}). 
The example in Figure~\ref{fig:perstep_comparison} shows improved pose and action details emerging during intermediate sampling and persisting in the final image.

\subsection{Ablation Studies}
\label{sec:ablations}

\textbf{Ablation setup.}
All ablations use Z-Image-Turbo and the same training subset, reporting GSB proportions and NetWin against the frozen baseline on 574 held-out prompts with matched seeds. Unless otherwise stated, we use corrected-rollout Flow-Bridge supervision, previous-velocity conditioning, and one corrector shared across $\mathcal{S}=\{2,3,4\}$, trained with Smooth L1 loss without backpropagation through rollout updates.

\begin{figure*}[t]
    \centering
    \includegraphics[width=\textwidth]{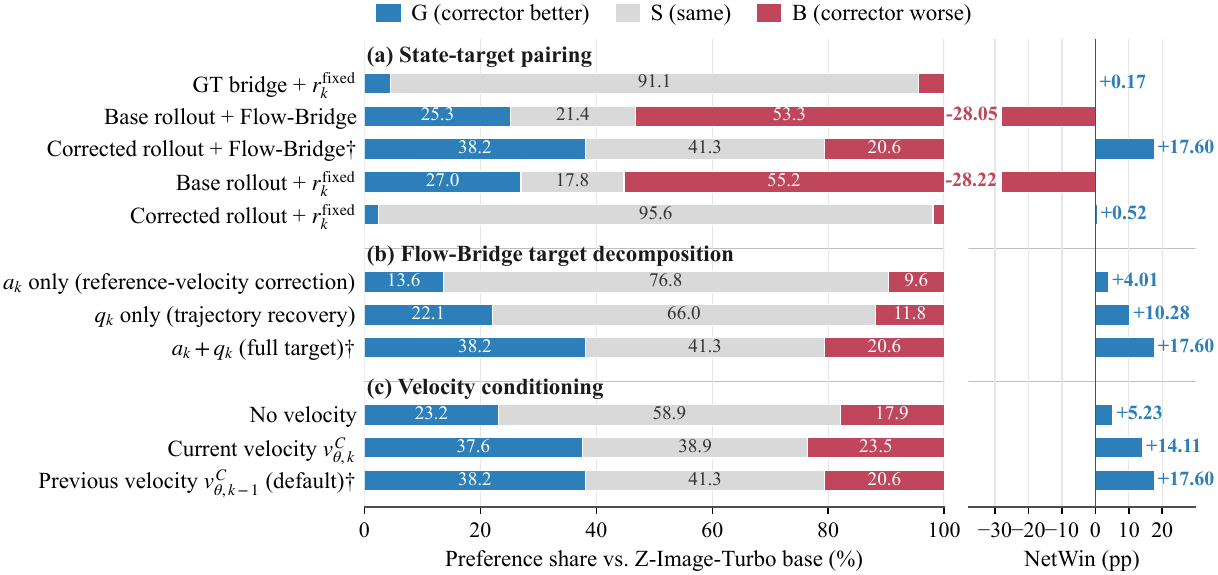}
    \caption{Ablations on Z-Image-Turbo: (a) state--target pairing, (b) Flow-Bridge target decomposition, and (c) velocity conditioning. We report GSB proportions (\%, left) and NetWin (pp, right) against frozen Z-Image-Turbo. $\dagger$ marks the full-method result repeated for reference.}
    \label{fig:gsb_ablations}
\end{figure*}

\textbf{State--target pairing.}
We test whether effective correction requires both corrected-rollout states and targets recomputed at those states (Figure~\ref{fig:gsb_ablations}(a)). With targets computed at each respective state, training on GT-bridge states yields little improvement, base-rollout training degrades performance, and corrected-rollout training produces the strongest gains. To separate rollout choice from target recomputation, we compare base and corrected rollouts using the same fixed target
$r_k^{\mathrm{fixed}}=\operatorname{sg}[v^{\mathrm{FB}}-v_\theta(b_k,\sigma_k,c)]$,
obtained by evaluating Eq.~(\ref{eq:bridge_residual}) at $b_k$; cached velocities come from the respective rollouts. With this fixed target, corrected rollouts largely remove degradation but yield little improvement over the baseline. Recomputing targets at corrected states then produces substantial gains, supporting the joint use of corrector-induced states and state-dependent supervision.

\textbf{Flow-Bridge target decomposition.}
Following Eq.~(\ref{eq:bridge_decomposition}), we compare reference-velocity correction
$a_k=v^{\mathrm{FB}}-v_{\theta,k}^C$,
trajectory recovery
$q_k=(b_k-z_k^C)/h_k$,
and their sum under identical corrected-rollout training, with all targets detached. Both components improve NetWin individually, while the full target outperforms either alone (Figure~\ref{fig:gsb_ablations}(b)), supporting their complementary roles. Combining them increases both win and loss rates, with the larger rise in wins yielding the highest NetWin.

\textbf{Velocity conditioning.}
We compare zeroed velocity channels, current velocity $v_{\theta,k}^C$, and cached previous velocity $v_{\theta,k-1}^C$, holding the remaining configuration fixed (Figure~\ref{fig:gsb_ablations}(c)). Both velocity-conditioned variants outperform zeroed conditioning. Previous velocity achieves the highest observed NetWin, with a similar win rate and a lower loss rate than current velocity. It also removes dependence on the current backbone output, enabling parallel execution.

\begin{figure}[t]
    \centering
    \begin{subfigure}[t]{0.4\linewidth}
        \centering
        \includegraphics[width=\linewidth]{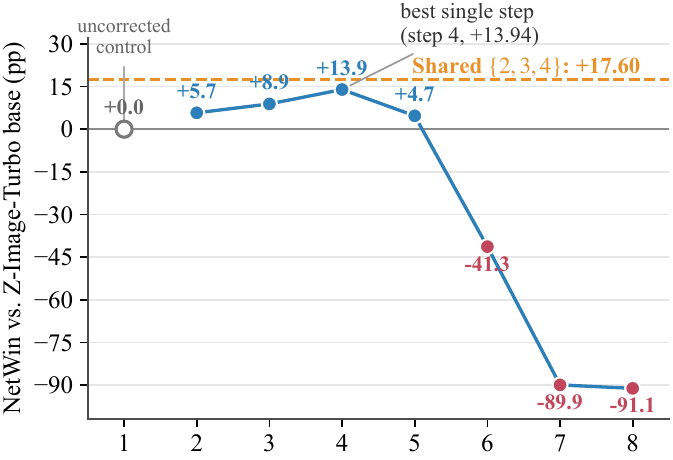}
        \caption{Single-step correction.}
        \label{fig:gsb_single_step}
    \end{subfigure}
    \hfill
    \begin{subfigure}[t]{0.55\linewidth}
        \centering
        \includegraphics[width=\linewidth]{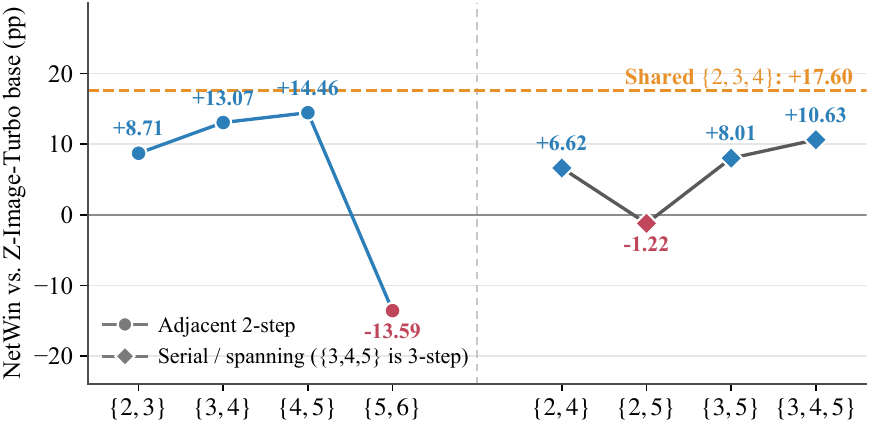}
        \caption{Multi-step correction.}
        \label{fig:gsb_two_step}
    \end{subfigure}
    \caption{Active-step ablations under eight-step Z-Image-Turbo sampling: (a) single-step and (b) multi-step correction. NetWin is evaluated against the frozen baseline on 574 held-out prompts with matched seeds. The orange dashed line denotes the default $\mathcal{S}=\{2,3,4\}$ configuration (+17.60 pp).}
    \label{fig:gsb_ablations_steps}
    \vspace{-3mm}
\end{figure}

\textbf{Active-step configurations.}
Under the same eight-step schedule, the default $\mathcal{S}=\{2,3,4\}$ achieves the highest NetWin among tested configurations, exceeding the best single-step ($\{4\}$) and two-step ($\{4,5\}$) variants (Figure~\ref{fig:gsb_ablations_steps}). More correction steps do not necessarily help: $\{3,4,5\}$ underperforms $\{4,5\}$, and late-step correction can severely degrade performance. This may reflect fewer remaining updates to accommodate correction-induced deviations, supporting intervention at selected early steps.

\section{Conclusion}
\label{sec:conclusion}

We introduced a lightweight parallel velocity corrector trained with on-trajectory Flow-Bridge supervision, preserving the frozen generator's parameters, sampling schedule, and generator forward-pass count. With fewer than 20M trainable parameters, it achieves +17.60 pp target-domain NetWin on Z-Image-Turbo with minimal inference overhead. 

Very short schedules, such as four-step generation, may leave insufficient updates after correction to complete denoising. Offline targets from the proprietary GPT-Image-2 editor may also introduce biases and limit reproducibility. Future work will explore schedule-aware correction across few-step generators and reproducible supervision from open models.

\section*{AI Use Statement}

We used generative AI tools for data construction, automatic evaluation, and manuscript preparation. For data construction, \texttt{claude-opus-4-8} assisted in expanding structured scene specifications into prompts and checking their consistency, atomic requirements, and subject assignments. GPT-Image-2 was used to edit frozen-generator outputs to construct offline target images for training. For GSB evaluation, \texttt{gpt-5-chat-0807} extracted visual requirements, and \texttt{claude-opus-4-8} assessed requirement satisfaction and severe structural failures. These uses are described in the data construction and evaluation sections and their corresponding appendices. We also used LLMs to assist with translation, language polishing, and refinement of the manuscript's organization and presentation.

The authors reviewed the AI-assisted manuscript and checked that its technical descriptions and reported results were consistent with the implemented methods and experimental records. AI-generated data and automatic judgments were processed using the checking and evaluation procedures described in the paper. The authors take responsibility for the final content of this work, including all text, claims, and artifacts produced with the assistance of generative AI.

\bibliography{iclr2027_conference}
\bibliographystyle{iclr2027_conference}

\clearpage
\appendix

\section{Additional Experimental Details}
\label{app:experimental_details}

\subsection{Data Construction and Overlap Checks}
\label{app:data_details}

We construct approximately 4K compositional captions by sampling structured scene specifications from curated pools of subjects, scenes, and actions under predefined constraints, and using \texttt{claude-opus-4-8}~\citep{anthropic2026opus48} to expand them into descriptions.
The descriptions undergo consistency checks, verification of decomposed atomic requirements, and checks for ambiguous subject assignment, with revisions as needed, before being formatted into final image-generation prompts.
The resulting captions cover multi-subject scenes, human actions, interactions, spatial relations, counting, and attribute binding. 

For each caption, we sample an initial noise $\epsilon$ and generate an image with the frozen backbone. The generated image and caption are provided to the GPT-Image-2-based editing model \texttt{gpt-image-2-0421-global-edit}~\citep{openai_gpt_image2}. The edited image serves as the supervision target, while the original caption and initial noise are retained to form the training tuple $(c,\epsilon,x^\star)$. The editor is used only for offline target construction and is not queried during corrector training or inference.

We reserve 10\% of the caption pool for validation and use the remainder for training.
Final target-domain evaluation uses 574 held-out prompts.
Exact string matching after whitespace normalization confirms that these evaluation prompts are unique and do not overlap with either the training or validation set.

\subsection{Architecture, Optimization, and Baselines}
\label{app:implementation_details}

\paragraph{Corrector architecture.} The S3-DiT corrector contains six Transformer blocks with hidden dimension 384, three attention heads, head dimension 128, and patch size two, totaling approximately 18.6M trainable parameters. A single corrector is shared across $\mathcal{S}=\{2,3,4\}$ of the frozen eight-step Z-Image-Turbo generator. It reuses the generator's precomputed text token embeddings without an additional text-encoding forward pass, and the generator and corrector branches execute in parallel at active steps.

\paragraph{Corrector optimization.} We train only the corrector using AdamW with learning rate $2\times10^{-4}$, weight decay $10^{-2}$, effective batch size 64, and gradient-norm clipping at 1.0. Training uses BF16 precision, a per-GPU micro-batch size of one, gradient checkpointing, and the Smooth L1 residual objective with $\beta=2$. The backbone, text encoder, and VAE remain frozen, and rollout states and cached velocities are detached between successive sampling steps, as described in Section~\ref{sec:shared_multistep}.

\paragraph{Baseline configuration.} Our implementation uses DiffSynth-Studio~\citep{diffsynthstudio}. The three LoRA baselines use rank $r=8$, scaling factor $\alpha=8$, and zero dropout, with adapters applied to the attention query, key, value, and output projections and the three feed-forward projections across the 30 main, two noise-refiner, and two context-refiner blocks. Each has 19,845,120 trainable student parameters; the original backbone weights, text encoder, and VAE remain frozen. All four trainable baselines use AdamW with a configured learning rate of $10^{-5}$, weight decay $10^{-2}$, and effective batch size 64, measured in preference pairs for DPO+LoRA.

\paragraph{SFT+LoRA.} We train the adapters by flow matching on paired image--text data. Given a target latent $x_0$ and Gaussian noise $\epsilon$, we construct $x_\sigma=(1-\sigma)x_0+\sigma\epsilon$ and regress the velocity target $\epsilon-x_0$. Training samples an index uniformly from a 1,000-point noise grid with shift three and uses the implementation's normalized Gaussian-envelope loss weighting.

\paragraph{DPO+LoRA.} We perform offline diffusion preference optimization with target images as chosen samples and fixed eight-step frozen-Turbo generations from the same prompts as rejected samples. Chosen and rejected samples share the same noise and noise level, and the frozen reference predictions are obtained by disabling the LoRA adapters. We use preference coefficient $\beta_{\mathrm{DPO}}=2000$ and select the checkpoint with the lowest validation DPO loss on a prompt-disjoint holdout.

\paragraph{D-OPSD.} We train student LoRA adapters along eight-step student-generated trajectories by matching the denoised latent estimates of an EMA teacher, with trajectory states detached between steps. The teacher is conditioned on the prompt and target image through a frozen Qwen3-VL-4B-Instruct model, and its adapters are updated with EMA decay 0.9999. Teacher parameters are excluded from the trainable parameter count. Inference uses only the student and the original text encoder.

\paragraph{DiffCon.} We train an independent controller with eighteen Transformer blocks, hidden dimension 256, two attention heads, and 20,206,401 trainable parameters (20.21M). It receives the current noisy latent and the frozen backbone's denoised estimate, together with text and timestep conditioning, and predicts a scalar $z$ and residual field $h$ to form $v_{\mathrm{DiffCon}}=v_{\mathrm{base}}+(1-z)h-zv_{\mathrm{base}}$. Its output projection is zero-initialized. Training uses flow-matching mean squared error on noised target latents, sampling uniformly from the eight inference noise levels. At inference, the controller is evaluated after the backbone at every sampling step.

\begin{figure}[t]
    \centering
    \includegraphics[width=\linewidth]{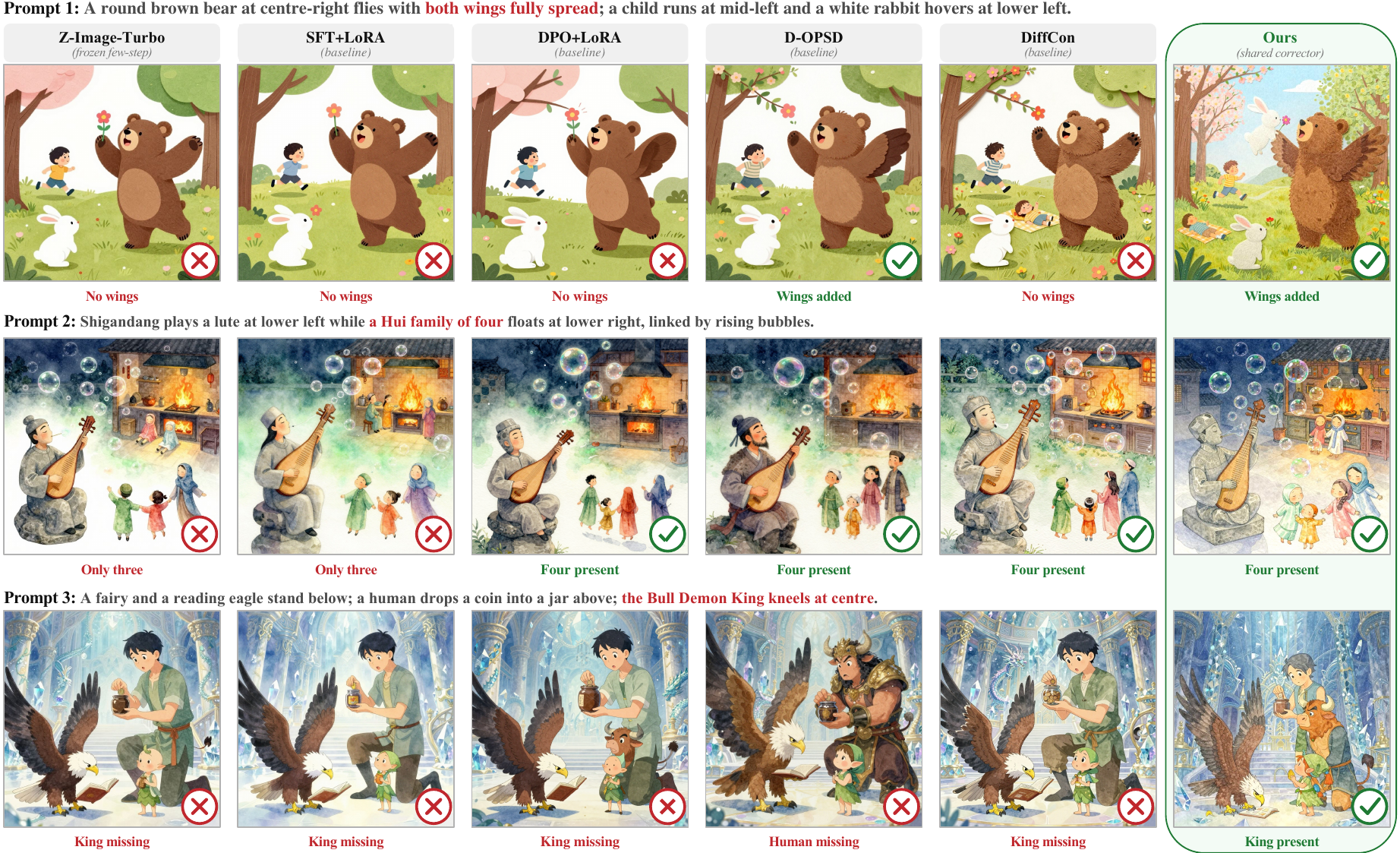}
    \caption{Additional qualitative comparisons of compositional instruction following in few-step text-to-image generation, supplementing Fig.~\ref{fig:results}.}
    \label{fig:motivation_additional}
    \vspace{-6mm}
\end{figure}

\subsection{GSB Judging Details and Repeatability}
\label{app:gsb_evaluation_details}

Evaluation consists of two stages. First, \texttt{gpt-5-chat-0807}~\citep{openai_gpt5_chat} decomposes each prompt into independently verifiable visual requirements, annotated by importance (core or detail) and dimension. Second, the visual judge \texttt{claude-opus-4-8}~\citep{anthropic2026opus48} compares each image pair requirement by requirement and checks for severe structural failures. Images are resized to a longest side of 768 pixels before judging.

Each pair is evaluated in both image orders. After mapping judgments back to model identities, we retain a directional requirement preference if at least one order supports it and neither order supports the opposite direction. Sample-level outcomes are determined lexicographically by severe-failure counts, net core-requirement preferences, and net detail-requirement preferences. Atmosphere requirements are recorded but excluded from the decision. 

For each judging run, NetWin is computed using $\mathrm{NetWin}=100(G-B)/N$ . Ties remain in the denominator, so the metric reflects both the direction of preferences and the prevalence of pairs without a decisive difference.

To assess automatic-judging stability, we perform five independent re-evaluations of the same fixed image pairs, applying the image-order reversal and aggregation protocol in each run. The standard deviation of the resulting run-level NetWin scores is below 1 percentage point. This measures judging repeatability on fixed outputs. GSB judgments are used only for evaluation and never as a corrector-training signal.

\subsection{T2I-CompBench++ Evaluation Details}
\label{app:compbench_evaluation_details}

We evaluate compositional generalization on T2I-CompBench++~\citep{huang2025t2icompbenchpp} using its official automatic evaluators: disentangled BLIP-VQA based on BLIP~\citep{li2022blip} for color, shape, and texture binding; UniDet-based metrics~\citep{zhou2022simple} for 2D/3D spatial relationships and numeracy; CLIPScore~\citep{hessel2021clipscore} for non-spatial relationships; and 3-in-1 for complex compositions.

We use the complete official test prompt set, comprising 300 prompts in each of eight categories. Our evaluation differs from the official protocol only in the number of generated images per prompt: we generate four images (seeds 0--3) instead of ten. We report the mean score over all 1,200 generated images in each category.

\subsection{Additional Qualitative Comparisons}
\label{app:additional_qualitative}

Figure~\ref{fig:motivation_additional} supplements Fig.~\ref{fig:results} with additional qualitative comparisons of compositional instruction following.

\end{document}